\documentclass[runningheads]{llncs}

\usepackage{eccv}

\usepackage{eccvabbrv}

\usepackage{graphicx}
\usepackage{booktabs}
\usepackage{algorithm}
\usepackage{algpseudocode}
\usepackage{algcompatible}
\usepackage{array}

\usepackage[accsupp]{axessibility}  

\usepackage{hyperref}

\usepackage{orcidlink}
\newcommand{\firstone}[1]{\colorbox{red!15}{#1}}
\newcommand{\secondone}[1]{\colorbox{blue!15}{#1}}

\begin{document}

\title{Vision Transformers with Natural Language Semantics} 

\titlerunning{Vision Transformers with Natural Language Semantics}
\author{
Young Kyung Kim$^{1}$,
J. Mat\'ias Di Martino$^{1}$,
Guillermo Sapiro$^{1,2}$\\
[2mm]
$^1$~Duke University  \quad $^2$~Apple 
}
\author{Young Kyung Kim\inst{1}\orcidlink{0000-0002-8769-6092} \and
J. Mat\'ias Di Martino\inst{1}\orcidlink{0000-0002-4679-8666} \and
Guillermo Sapiro\inst{1,2}\orcidlink{0000-0001-9190-6964}}

\authorrunning{Y.~Kim et al.}

\institute{Duke University, Durham, NC, 27708, USA
\email{\{yk206,matias.di.martino,guillermo.sapiro\}@duke.edu}\and
Apple, Cupertino, CA, 95014, USA}

\maketitle

\begin{abstract}
Tokens or patches within Vision Transformers (ViT) lack essential semantic information, unlike their counterparts in natural language processing (NLP). Typically, ViT tokens are associated with rectangular image patches that lack specific semantic context, making interpretation difficult and failing to effectively encapsulate information. We introduce a novel transformer model, Semantic Vision Transformers (sViT), which leverages recent progress on segmentation models to design novel tokenizer strategies. sViT effectively harnesses semantic information, creating an inductive bias reminiscent of convolutional neural networks while capturing global dependencies and contextual information within images that are characteristic of transformers. Through validation using real datasets, sViT demonstrates superiority over ViT, requiring less training data while maintaining similar or superior performance. Furthermore, sViT demonstrates significant superiority in out-of-distribution generalization and robustness to natural distribution shifts, attributed to its scale invariance semantic characteristic. Notably, the use of semantic tokens significantly enhances the model's interpretability. Lastly, the proposed paradigm facilitates the introduction of new and powerful augmentation techniques at the token (or segment) level, increasing training data diversity and generalization capabilities. Just as sentences are made of words, images are formed by semantic objects; our proposed methodology leverages recent progress in object segmentation and takes an important and natural step toward interpretable and robust vision transformers. 
\end{abstract}

\section{Introduction}
\label{sec:intro}
Transformer has achieved a significant breakthrough in Natural Language Processing (NLP)  \cite{vaswani2017attention, devlin2018bert, brown2020language}, leading to its widespread adoption in Computer Vision (CV) where it has quickly made a similar impact  \cite{dosovitskiy2020image, liu2021swin, bao2021beit}. In contrast to the previously dominant Convolutional Neural Networks (CNN)  \cite{lecun1989backpropagation, krizhevsky2012imagenet, he2016deep}, transformer-based vision models gained popularity over standard CNN due to their ability to capture non-local dependencies and contextual information within images. Consequently, these models have attained the State-Of-The-Art (SOTA) performance in various tasks, including classification, object detection, segmentation, and image captioning  \cite{zhai2022scaling, fang2023eva, li2023mask, radford2021learning}.

\begin{figure}[t]
  \centering
   \includegraphics[width=0.40\linewidth]{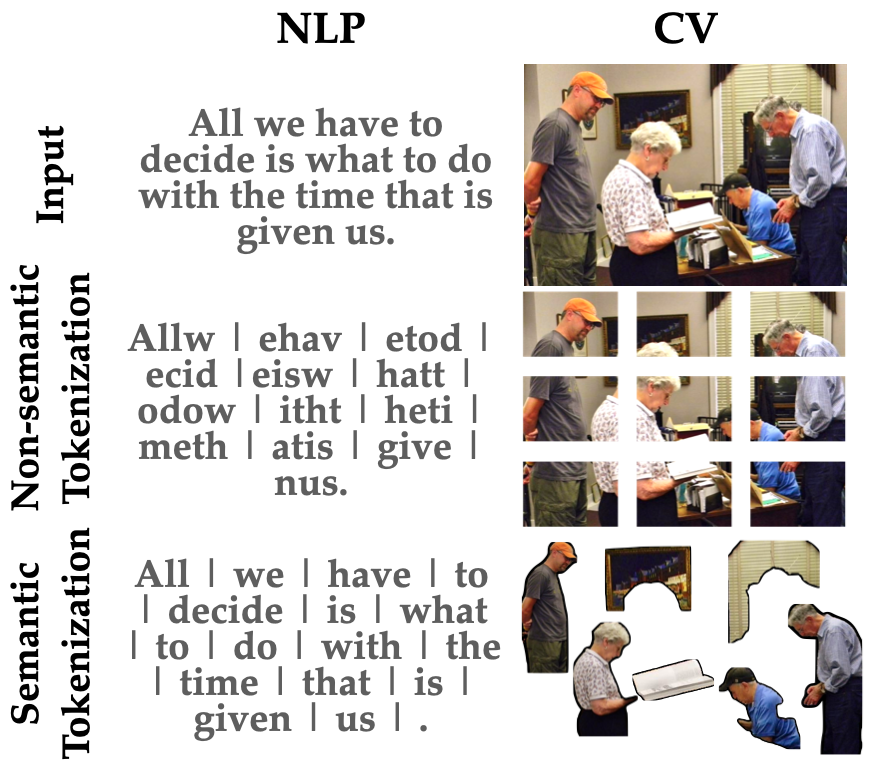}

   \caption{Example of non-semantic and semantic tokenization for NLP and CV. Tokens from non-semantic tokenization are hard to explain and understand, while it is easier to explain and understand tokens from semantic tokenization for both NLP and CV.}
   \label{fig:intro}
\end{figure}

Despite the analogous impact of transformers in CV and NLP, the way they are applied differs, particularly in the aspect of tokenization. The key distinction lies in the semantic information carried by tokens. In NLP, tokens carry high-semantic information as we understand the meaning of individual words or tokens. In contrast, tokens in CV possess lower semantic information since images are represented as a collection of regular patches. Following the analogy with NLP methods, this is equivalent to implementing NLP models breaking up sentences every fixed number of characters (instead of words), as illustrated in \cref{fig:intro}. When examining the non-semantic tokenization, as depicted in the second row of \cref{fig:intro}, it can be challenging to discern their meaning, even in the context of NLP. In contrast, semantic tokenization, shown in the third row, provides a clear understanding of the meaning of each token and leads to attention mechanisms with semantically-rich individual elements.

\begin{figure}[t]
  \centering
   \includegraphics[width=0.65\linewidth]{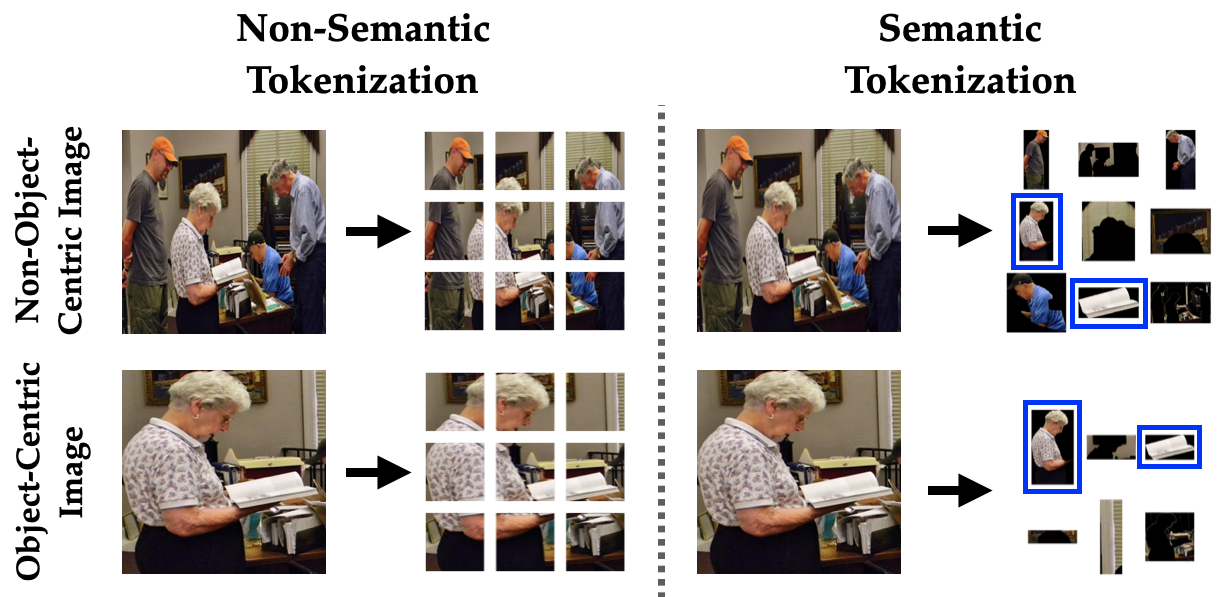}

   \caption{The example illustrates the differences between non-semantic and semantic tokenization on non-object-centric and object-centric images. As shown in the left column under non-semantic tokenization, patches containing women differ between non-object-centric and object-centric images, whereas they are similar or identical to when semantic tokenization is applied. This consistency is achievable because semantic tokenization segments based on object presence rather than uniform patch size.}
   \label{fig:invar}
\end{figure}

There are numerous advantages to utilizing semantic tokenization methods as here proposed for CV. Firstly, this form of tokenization reduces the search space by incorporating semantic prior knowledge into the model. For example, it provides information about the grammatical structure for NLP models and similar contextual information could be leveraged for CV models. In particular, it reduces the number of tokens required as input for the model, further decreasing the search space, which is especially useful for small training datasets (empirical examples and evidence are provided in \cref{sec:Experiments}). Secondly, this semantic tokenization enables the vision model to be scale invariance, enhancing robustness for transfer-learning on out-of-distribution datasets, particularly transfer learning between object-centric and non-object-centric datasets. As illustrated in \cref{fig:invar}, patches on a woman differ significantly between non-object-centric and object-centric data when using non-semantic tokenization. However, with semantic tokenization, the patches on the woman are consistent across both datasets. Therefore, vision models utilizing semantic tokenization are expected to generalize better, even when the scale of objects varies across different datasets.

Thirdly, semantic tokenization enhances the interpretability of the model. Most current interpretation methods assume that the model and humans perceive semantics in the same way. However, as noted by Hoffmann et al. \cite{hoffmann2021looks}, this assumption does not hold, as some highlighted parts for interpretation lack any semantic meaning. Our new method, trained to separate images into humanly interpretable semantic tokens, overcomes such limitations (see experimental examples in \cref{sec:Experiments}). Fourthly, this form of tokenization expands the possibilities for augmentation. Traditionally, most augmentations have been applied to entire images. However, when applied to individual semantic tokens as in this work, it significantly increases diversity. Lastly, the semantic tokenizer here introduced captures multi-scale information by construction, and provides information on positional embedding. Presently, all transformer-based vision models take the patch order as the input for positional embedding. However, by utilizing semantic tokenization, we use the position and pixel size of the extracted segment in an image as input for positional embedding. This provides rich multi-scale and spatial information and provides the opportunity for extremely efficient and rich positional data augmentation as we describe and discuss in \cref{sec:method}. 

Despite its clear advantages, semantic tokenization is not widely used in computer vision due to its greater complexity compared to NLP. Unlike NLP, where spaces separate words, CV lacks a universal separator, leading to varied and domain-specific segmentation methods. These complexities have made semantic tokenization a distinct task in CV. In this paper, we address this and introduce Semantic Vision Transformer (sViT), leveraging recent advancements in segmentation models. To the best of our knowledge, this is the first approach that utilizes semantic image tokenization in the context of transformer-based vision models.

Our main contributions are: i) We identify limitations of transformers in CV compared to NLP, as CV utilizes non-semantic patches as tokens instead of semantic ones as NLP does, and we provide a practical alternative solution; (ii) We demonstrate our proposed method is more expressive compared to the standard (non-semantic) approach, especially in out-of-distribution generalization and robustness to natural distribution shifts; (iii) We propose a new paradigm for data augmentation, applying augmentation on each semantic segment instead of on the entire image, thereby increasing diversity. We also increase the model's robustness with respect to changes in the positions and sizes of the semantic elements; and (iv) While patches in standard vision transformers and highlighted regions from interpretation methods on vision model are not necessarily explainable, our new semantic tokenization method is improving interpretability for learned vision model. We describe the proposed methods in \cref{sec:method}. We present empirical evidence and interpretability results in \cref{sec:Experiments}. We discuss related work in \cref{sec:rw}. Finally, we provide conclusions in \cref{sec:conclusion}.

\section{Methods}
\label{sec:method}
\subsection{sViT}
\label{sec:sViT}

For constructing a semantic tokenization, we exploit the Segment Anything Model (SAM) \cite{kirillov2023segment}, which has been trained on $11$ million images for segmentation tasks. Each segment is resized to $16\times16$ or $32\times32$ pixels, and we utilize convolutional layers of matching sizes to flatten them into vector embeddings, following an approach similar to the original Vision Transformer (ViT). Instead of using the position or order of patches for positional encoding, we employ the horizontal and vertical $(x,y)$ image coordinates of the bounding box and the pixel size of each segmentation as provided by SAM. 


The coordinates of the bounding box provide relative positional and contextual information about objects in the image, which is crucial for comprehending the image content. For instance, an image with a person on top of a horse conveys a different meaning than an image with a person beside a horse, indicating the potential actions of riding or tending to the horse, respectively. The segment size information is also valuable, serving as a signal for the model to identify which objects may be essential for understanding the image. This information is added to the tokens, and a detailed explanation of the process is summarized in the supplementary material.

\begin{figure*}[t]
  \centering
   \includegraphics[width=1\linewidth]{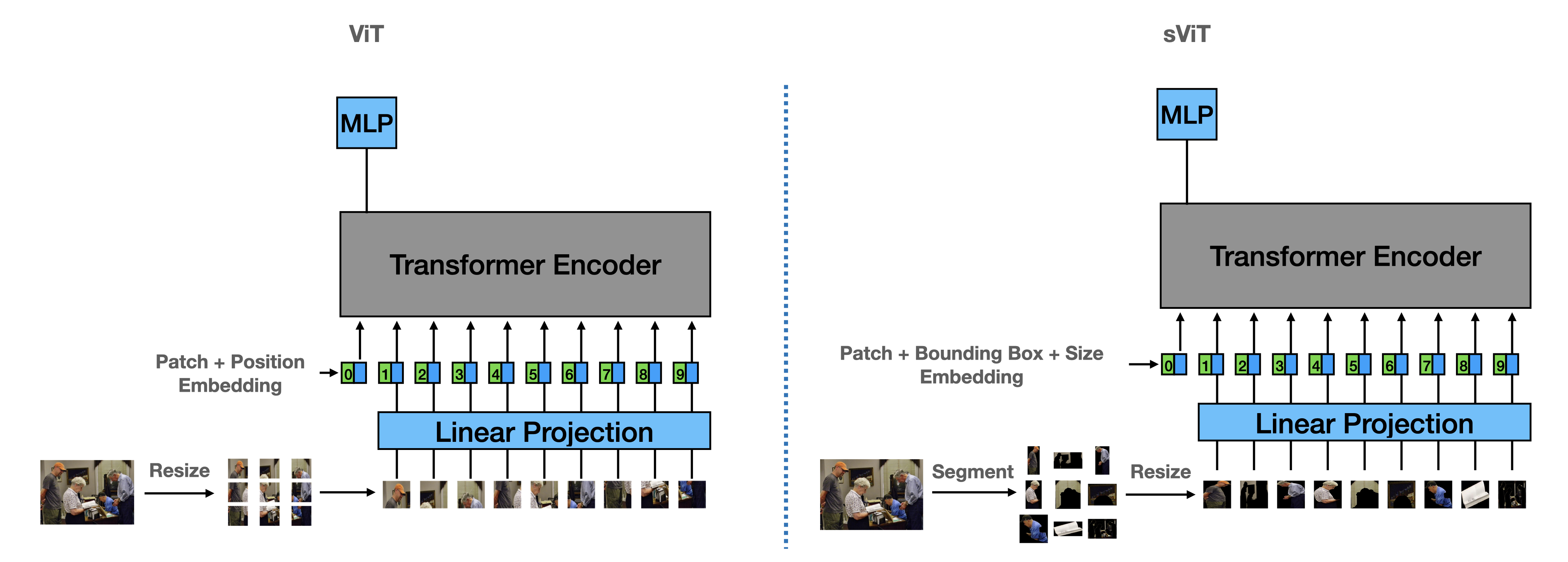}
   \caption{Architectures of ViT and sViT. ViT first resizes an image to a fixed size and divides it into equal-sized patches, whereas sViT divides an image at the semantic segments level and resizes the segments into equal-sized patches. ViT incorporates positional embedding mapped from the order of patches, while sViT includes positional embedding mapped from the coordinates of the semantic bounding box and the sizes of the segments. Outside of these two fundamental differences, the transformers architecture remains the same.}
   \label{fig:scratch_mod}
\end{figure*}

We utilized the basic architecture of the original ViT model, designed to process $196$ tokens as input. In contrast to the original ViT, our model accommodates varying token lengths as input, given that SAM produces a different number of segments depending on the input image. To handle this variability, we introduce a token referred to as the `background token,' which represents the remaining pixels after segmenting all objects in the image. To designate this as a background token, we assign $-1$ values to the $x$ and $y$ coordinates of the bounding box, similar to the concept of special tokens in language models. For images with more than $195$ segments, we utilize the first $195$ segments as tokens and assign the remaining segments to the background token. This approach ensures that no elements of the original image are discarded. This model is illustrated and compared to the original ViT in \cref{fig:scratch_mod}. 

\subsection{Data Augmentation}
\label{sec:aug}

\begin{figure}[h]
  \centering
   \includegraphics[width=0.75\linewidth]{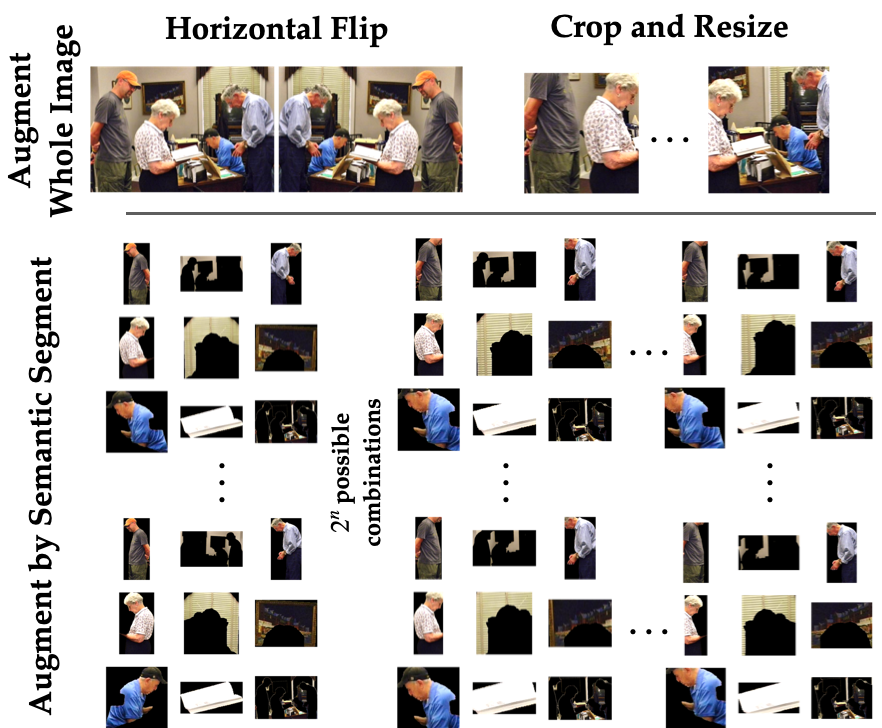}
   \caption{Example of augmented images at the semantic segment level and the entire image level. Top-row images are data autmentation results using horizontal flip, cropping, and resizing techniques on the entire image. Bottom-row images are data autmentation results using the same techniques but now at the semantic segment level.}
   \label{fig:aug}
\end{figure}

Data augmentation techniques have shown to lead to strong generalization performance. Most of these augmentations are applied to entire images, limiting their capability to encourage relative (semantic) component diversity for a given scene. For example, when we apply a horizontal flip to an entire image, all objects within the image are flipped and maintain their relative relations. To achieve a wider range of data augmentations from a single image, it is essential to apply augmentations at the segment level. Applying augmentations at the segment level also enhances crop and resize techniques. When we apply crop and resize techniques to an entire image, there is a risk of losing objects due to such cropping. For instance, in \cref{fig:aug}, focusing on cropping the female in the image might result in some objects being cropped out. However, when we apply cropping and resizing techniques at the segment level, we can retain all objects and create a greater variety of data augmentations. 

By performing augmentation at the token level, we further avoid the complex task of re-rendering a complete scene. Moreover, our new tokenizer introduces additional inputs for positional embedding, which are the positions and sizes of image segments. These inputs enable us to implement an innovative augmentation method, namely introducing noise into the positional and size inputs for each semantic token. Consider an image of a group of horses running on a plain. Minor variations in the relative positions of the horses or changes in their sizes might result from shifts in camera angle or distance. Introducing noise to the positional and size data for each semantic token helps the vision model generalize to such variations, a feature not addressed by current global data augmentation methods.

To integrate these concepts, the augmentation function for sViT accepts the following inputs: a list of segments, positional and size data for segments, the maximum percentage for augmentation sampling, and functions for augmentations, as detailed in \cref{alg:aug_seg}. Rather than applying augmentations to the same number of segments in each epoch, we introduce variability by sampling a percentage for augmentation. This approach enhances diversity by exposing the image to varying levels of augmentation intensity during training. 

\begin{algorithm}[h]
   \caption{Augmentation by Segment Level}
   \label{alg:aug_seg}
\begin{algorithmic}
    \STATE {{\bfseries Output:} List of segments: $list\_seg$, List of bounding box and size: $list\_bb\_size$}
   \STATE {{\bfseries Input:} List of segments: $list\_seg$, List of bounding box and size: $list\_bb\_size$, Maximum percent to sample: $max\_perc$, Horizontal flip: $hf$, Crop and resize: $cr$, Gaussian noise: $gn$, $n$ unique sampler from list $l$: $rc(n,l)$, Uniform distribution: $U$}
   \\\hrulefill
   \STATE {$perc\_samp \sim U(0, max\_perc)$}
   \STATE {$num\_samp = len(list\_seg) \cdot perc\_samp$}
   \STATE {$list\_index = rand(num\_samp,list\_seg)$}
   \FOR{$i=1$ {\bfseries to} $list\_index$}
   \STATE {$list\_seg[i] = hf(list\_seg[i])$}
   \STATE {$list\_seg[i] = cr(list\_seg[i])$}
   \STATE {$list\_bb\_size[i] = gn(list\_bb\_size[i])$}
   \ENDFOR
   \STATE {Return $list\_seg$, $list\_bb\_size$}
\end{algorithmic}
\end{algorithm}

\subsection{Enhancing Model Interpretability}
\label{sec:int}
Similar to other gradient-based methods, we utilize gradient propagation to identify the critical semantic segments in an image for predicting specific classes. In contrast with existing methods, instead of relying solely on gradient information from the last layer of the vision model, we use gradient information on the semantic (and interpretable) tokens. The sViT tokenizer, a segmentation model, is trained in a supervised manner to learn how humans separate an image into distinct and understandable parts. 

To determine the gradient-based importance level ($I^{c}_{i}$) for each token $i$ when predicting class $c$, the initial step involves computing the gradient of the predicting class $c$ ($y^{c}$), with respect to the embedding of the semantic token $i$ ($T_{i}$), $\dfrac{\partial y^{c}}{\partial T_{i}}$. The subsequent process includes element-wise multiplication of the gradient with the token embedding, followed by computing the average of the multiplied tensor. Similar to other gradient-based methods, we applied the $ReLU$ function to the computed average to convert all negative values to zero. The complete equation becomes $I^{c}_{i} = ReLU\left(\dfrac{1}{N} \sum_{j} \left(\dfrac{ \partial y^{c}}{\partial T_{i,j}} \cdot T_{i,j}\right)\right)$. 


\section{Experiments}
\label{sec:Experiments}
We now assess how the proposed sViT combines the dual advantages of possessing an inductive bias and capturing global dependencies and contextual information among segments of an image, and compare its performance in both small and large datasets to that of ViT. Additionally, demonstrates how the proposed sViT's inductive bias contributes to transfer learning, particularly in small datasets. Our evaluation also demonstrates the effectiveness of the proposed semantic data augmentation and the natural explainability of sViT.

\subsection{Setup}
\subsubsection{Datasets}
We use three scene recognition datasets: Places365 (Places) \cite{zhou2017places}, MIT Indoor Scenes (MIT 67) \cite{quattoni2009recognizing}, and SUN397 (SUN) \cite{xiao2010sun}. We specifically selected these datasets due to their complexity and non-object-centric nature. For out-of-distribution generalization, we use eight object-centric datasets: FGCV Aircraft (Aircraft) \cite{maji2013fine}, Stanford Cars (Cars) \cite{krause20133d}, Caltech 101 (Caltech) \cite{fei2004learning}, ImageNet (INet) \cite{deng2009imagenet}, Describable Textures Dataset (DTD) \cite{cimpoi14describing}, Flower Dataset (Flower) \cite{Nilsback08}, Food 101 (Food) \cite{bossard14}, and Pet Dataset (Pet) \cite{parkhi2012cats}.

For evaluating robustness to natural distribution shift on INet, we used four evaluation datasets: ImageNet V2 (INet-V2) \cite{recht2019imagenet}, ImageNet-A (INet-A) \cite{hendrycks2021natural}, ImageNet-Renditions (INet-R) \cite{hendrycks2021many}, and ImageNet-Sketch (INet-S) \cite{wang2019learning}. More details on training and evaluating datasets are in the supplementary material.

In our preprocessing phase, we employed SAM to isolate segments from each image. These segments are then resized to either $16\times16$ or $32\times32$ pixel dimensions and organized into a blank $224\times224$ or $448\times448$ pixel image, respectively. Additionally, for numerical stability, we normalized the $x$ and $y$ coordinates of the bounding box, as well as the size of segments for each image. For data-loading and augmentation, we used the FFCV library \cite{leclerc2023ffcv} and adapted its functionalities to segment-level data augmentation. 

\subsubsection{Training and Finetuning}
We trained the ViT model, and the proposed sViT with resolutions of $16\times16$ and $32\times32$ for each semantic token, which we refer to as sViT$16$ and sViT$32$, respectively. All models share the same architecture, comprising of $12$ heads and $12$ layers of transformers. For the ViT model, we applied augmentation using the cropping and resizing technique with horizontal flip. For sViT, we employed a maximum percentage of $25\%$ for sampling to augment, and cropping and resizing techniques with horizontal flip. For fine-tuning, we employed a linear evaluation protocol, in which only the linear classifier is updated while all other weights in the transformer are frozen. More details on parameters for training and fine-tuning are in the supplementary material.

\subsection{Results}
\label{Results}
\cref{tab:transfer} summarizes the performance comparison for ViT and sViT with and without data augmentation on non-object-centric datasets. Our proposed method sViT outperforms ViT for all the tested datasets. For example, our method outperforms ViT by at least two percent in accuracy when evaluated on a subset of Places dataset. This is a significant achievement, especially considering that sViT takes an average of 99 tokens, while ViT takes 196 tokens for each input image. This difference could be attributed to sViT having an inductive bias and prior knowledge implicitly extracted from SAM. Semantically meaningful tokenization (as in NLP) provides additional knowledge that reduces the search space, hence requiring less training data to achieve a certain performance level. For smaller datasets, the advantages of semantically rich tokens are even more evident, and our models significantly outperform ViT in MIT67 and SUN datasets.

\begin{table*}
\caption{Evaluation of ViT, sViT$16$, and sViT$32$ models on object-centric and non-object-centric datasets, with and without data augmentation (aug). In all datasets, sViT outperforms ViT. We highlighted \firstone{best} and \secondone{second-best} performances for each dataset.}
  \centering
  \resizebox{1\columnwidth}{!}{%
  \begin{tabular}{lcccc|ccccccccr}
  \toprule
   &  \multicolumn{4}{c}{non-object-centric dataset} & \multicolumn{8}{c}{object-centric dataset} \\
    \toprule
    Method      & Places   & Places & MIT67 & SUN & INet & Aircraft & Cars & Caltech  & DTD & Flower & Food               & Pet \\
                & ($25\%$) &        &       &     &      &          &       &        &        &    &        &        &     \\
    \midrule
    ViT & 32.19 & 46.93 &61.72 & 48.84 & \secondone{56.44} &51.94 & 40.20 & 68.78  & \secondone{50.95} & 50.46 & \secondone{63.36} & 35.07\\
    ViT aug & 33.20 & \secondone{47.64} & 64.78 & 50.59 & 54.78 & 48.75 & 54.81 & 63.86 & \firstone{53.77} & 47.69 & \firstone{71.31} &34.20\\
    \midrule
    sViT$16$ & 35.39 & 46.26 & 68.96 & 50.77 & 54.48 &76.99 & \secondone{72.90} & 76.03  & 48.40 & 46.06 & 57.51 & 28.00\\
    sViT$16$ aug & 37.42 & 46.72 & \secondone{70.75} & 51.03 & 54.68 & 77.01 & 72.46 & 76.36  & 48.88 & 37.78 & 58.14 & 26.44\\
    sViT$32$ & \secondone{37.47} & 47.02 & 70.59 & \secondone{51.06} & 56.35 & \firstone{79.97} & 71.85 & \secondone{76.85} & 48.08 & \secondone{75.45} & 59.18 & \firstone{51.50}\\
    sViT$32$ aug & \firstone{37.71} & \firstone{48.34} & \firstone{72.01} & \firstone{51.53}&  \firstone{56.81} & \secondone{79.23} & \firstone{73.60} & \firstone{77.07} & 44.78 & \firstone{75.72} & 59.45 & \secondone{50.65}\\
    \bottomrule
  \end{tabular}
  }
  \label{tab:transfer}
\end{table*}

As discussed for example in \cite{lee2021vision}, ViT is expected to learn from big data without any inductive bias. However, our sViT with a token size of 32, continues to outperform ViT with or without augmentation. This could be attributed to sViT's ability to capture global dependencies and long-range interactions among semantic segments. Augmentation by semantic segments has proven effective in improving the performance, especially in smaller datasets as shown in \cref{tab:transfer} for MIT67, SUN, and a subset of Places. Moreover, the token size also plays a crucial role, with larger token sizes in sViT outperforming smaller token sizes. This could be due to losing less information when resizing a large segment into smaller, fixed sizes. Employing both techniques together has shown optimal performance across datasets. See figures $2 \& 3$ in the supplementary material for additional examples. 

\cref{tab:transfer} also summarizes the performance comparison of ViT and the purposed sViT, both with and without data augmentation, across various object-centric datasets. Among eight datasets, sViT demonstrates a significant increase in accuracy, ranging from $10\%$ to $29\%$. This improvement is largely attributed to the model's scale invariance, as described in \cref{sec:intro}. The superior performance of sViT on the Aircraft, Cars, Flower, and Pet datasets can be further attributed to the prevalence of planes, cars, flowers, and pets in the Places dataset. Due to its exposure to a wide variety of those objects, combined with its scale invariance attribute, sViT effectively transfers its semantic knowledge to different datasets. Yet, sViT demonstrates inferior performance in the DTD and Food dataset. This may stem from its limited exposure to images of food and textures during pretraining, as the primary dataset used (Places) mainly contains scenery images, which likely include fewer instances of food or specific textures. This also aligns with marginal (less than one percent) improvement of sViT over ViT in INet, as out of 1000 labels, some of labels would not have exposure during training. Enhancing a model's generalization capabilities across different data types requiress pretraining on a broad and varied dataset.

There is increasing interest in vision and language models, which enables training on larger and diverse datasets \cite{radford2021learning}. In this context, sViT's out-of-distribution generalization becomes invaluable, as its semantic tokenization approach appears promising in narrowing the performance gap between models pretrained on object-centric and non-object-centric datasets, as shown in these result. This approach not only offers the benefits of cost and size but also encourages the model to learn the contextual relationships among objects within an image and allows the model to encounter a more diverse array of objects.

\subsection{Interpretability}
\label{sec:interpretability}

\begin{figure}[t]
  \centering
  \includegraphics[width=0.85\linewidth]{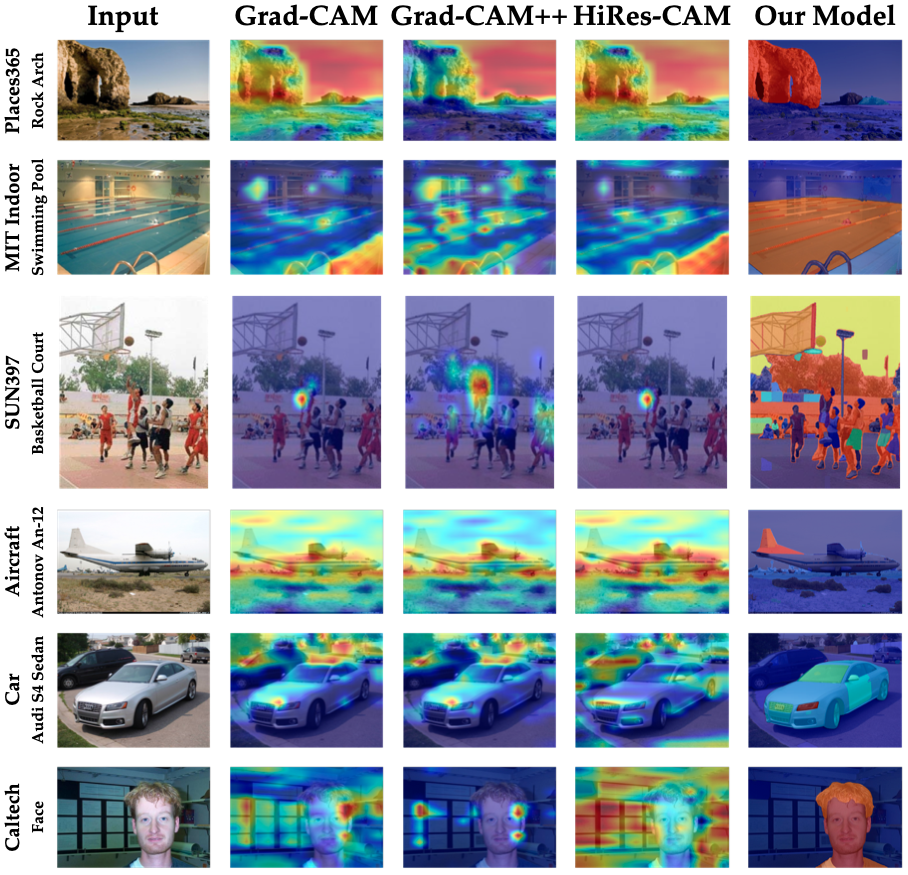}
   \caption{The first column shows the images used to evaluate the interpretability of ViT and sViT. Columns two to four represent the interpretable outcomes from applying Grad-CAM, Grad-CAM++, and HiRes-CAM on ViT, respectively. The fifth column shows the interpretable outcome for our method. The first three rows are from object-centric datasets and the last three rows are from non-object-centric datasets. The color coding indicates the level of importance. The importance escalates progressively through blue, green, yellow, and orange before reaching the peak at red. Only our proposed method consistently provides humanly interpretable and semantic results.}
   \label{fig:interpret}
\end{figure}

We examined and compared our approach with Grad-CAM, Grad-CAM++, and HiRes-CAM \cite{selvaraju2017grad,chattopadhay2018grad,draelos2020use}. Grad-CAM utilizes the gradient of the last layer of the model to highlight the important image regions for predictions, assuming that the last layer contains some semantic information. In contrast, Grad-CAM++ uses the second gradient, and HiRes-CAM computes element-wise multiplication between the gradient and feature map instead of multiplying the average of the gradient with the feature map. As shown in \cref{fig:interpret}, our proposed method is more interpretable. The primary reason for this lies in our model's use of SAM to segment an image into explainable tokens. Consequently, each highlighted segment is automatically interpretable. For instance, the arch is highlighted for the rock arch; multiple players and a basketball backboard are highlighted for the basketball court; and swimming lanes are highlighted in the swimming pool. In contrast, for other methods, it is challenging to semantically explain which part is being highlighted. Even when we examine the most interpretable example, such as Grad-CAM applied to the Aircraft dataset, it is difficult to discern whether the highlights indicate the aircraft or its background, as both are highlighted. Human have the tendency to use their bias to assume that the highlighted part represents some part of the aircraft as it is more intuitive to them. However, this will be an inaccurate interpretation as it discards the other highlighted parts. This challenge becomes even more noticeable in other examples, such as the face or swimming pool, where it is hard to describe the specific region the model is highlighting, obscuring the model's interpretation.

These limitations should be expected, as discussed by Locatello et al. \cite{locatello2019challenging}, disentangling representations into a few explanatory variables using unsupervised learning is challenging without a strong inductive bias. Grad-CAM and its variants lack disentanglement or inductive bias in the last layer to provide explanatory variables from the model. Therefore, the highlighted regions may appear random to human observers. This is evident in \cref{fig:interpret} (see figures $2 \& 3$ in the supplementary material for additional examples), as there is no exact algorithm that consistently outperforms the others among Grad-CAM-style methods. In comparison, sViT disentangles images into explainable semantic segments using supervised learning. Consequently, sViT is consistently more interpretable compared to Grad-CAM and its variants.

\begin{figure}[t]
  \centering
  \includegraphics[width=0.5\linewidth]{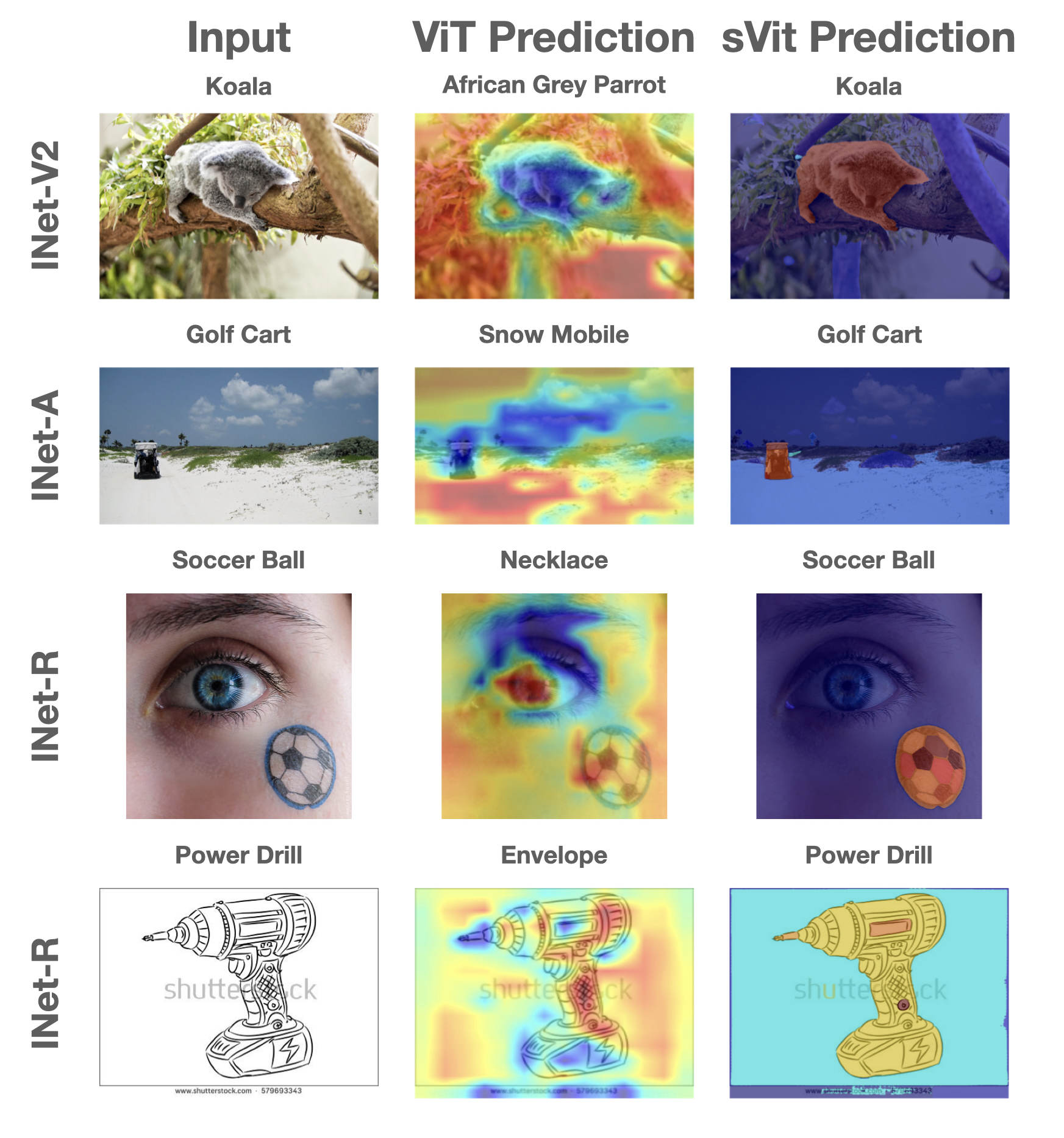}
   \caption{The first column shows the images used to evaluate the interpretability of ViT and sViT. Column two represent the interpretable outcomes from applying Grad-CAM on ViT. The third column shows the interpretable outcome for our method. The color coding indicates the level of importance. The importance escalates progressively through blue, green, yellow, and orange before reaching the peak at red.}
   \label{fig:robust}
\end{figure}

\subsection{Robustness on natural distribution shift}
\label{Robust}

sViT$32$ has shown superior robustness to natural distribution shifts when compared with ViT, achieving over two percent higher accuracy across all datasets. Notably, even sViT$16$, which generally underperforms relative to ViT on INet, outperforms ViT in three out of four evaluation datasets. This performance is especially significant in the INet-A, INet-S, and INet-R datasets, characterized by more pronounced distribution shifts—as evidenced by the more substantial performance decreases relative to INet-V2, on average. This resilience might be attributed to sViT's incorporation of semantic prior knowledge and the resizing of semantic segments. In \cref{fig:robust}, ViT appears to focus more on the background rather than the object of interest during prediction, especially when the object is small. This tendency in ViT, treating each pixel equally at the token level, contrasts with sViT, which considers semantic segments as input and assigns equal weights to segments rather than pixels at the token level. This approach, similar to how language models weigh words over characters, seems to direct the model's focus more towards objects of interest rather than the background with giving equal weights. This characteristic could explain why sViT exhibits increased robustness to natural distribution shifts. By prioritizing objects of interest, sViT maintains its robustness against background changes or minor perturbations to the object of interest.

\begin{table}
    \caption{Evaluation of robustness of models on natural distribution shift for ImageNet. We highlighted \firstone{best} and \secondone{second-best} performances for each dataset.}
  \centering
  \begin{tabular}{lc|ccccr}
    \toprule
    Method & INet & INet-V2 &    INet-A & INet-R  & INet-S \\
    \midrule
    ViT      & \secondone{56.44} & 44.29 & 1.51 & 12.79 & 10.27 \\
    ViT aug      & 54.78 & 42.85 & 1.29 & 10.53 & 8.05 \\
    sViT$16$ & 54.48 &	43.21 &	4.15 &	15.11 & 14.15  \\
    sViT$16$ aug & 54.68 &	43.90 &	4.03 &	15.05 & 13.81  \\
    sViT$32$ & 56.35 &	\secondone{45.75} &	\secondone{4.72} &	\secondone{15.80} &	\secondone{14.60}  \\
    sViT$32$ aug & \firstone{56.81} &	\firstone{46.27} &	\firstone{4.79} &	\firstone{16.44} &	\firstone{15.59}  \\
    \bottomrule
  \end{tabular}
  \label{tab:robust}
\end{table}


\section{Related Works}
\label{sec:rw}

Convolutional Neural Networks (CNN) have been the dominant image model for various tasks \cite{lecun1989backpropagation, krizhevsky2012imagenet, he2016deep}. With the emergence of transformers, they have gained significant attention, especially due to their state-of-the-art performance in various image tasks \cite{zhai2022scaling, fang2023eva, li2023mask, radford2021learning}. CNN possesses the inductive bias of locality and weight sharing, making them equivariant to translation and known to require less data for training. In contrast, ViT has the advantage of capturing global dependencies and long-range interactions among images patches. The here proposed sViT combines both advantages; it has an inductive bias using a semantic segmentation model, making the model equivariant to translation and requiring less data for training, capturing semantically rich interactions, and enhancing data-augmentation and interpretability capabilities of existing ViT approaches. 

Various research efforts have sought to enhance tokenizers for ViTs. To introduce an inductive bias into ViTs, several methods have been explored, including the use of overlapping patches, diversifying tokens through image augmentation, and employing a vector quantized encoder that outputs discrete embeddings \cite{yuan2021tokens, xu2023scale, mao2021discrete}. Additionally, many studies have aimed to reduce the number of tokens to enhance model efficiency during inference, e.g., \cite{liang2022not, yin2022vit}. Despite these efforts, little attention has been directed toward making tokens more semantically meaningful. By enhancing the semantic aspect of tokens, we have successfully added an inductive bias, introduced greater diversity in augmentations, and improved the interpretability of vision models. This resulted in overall improved performance.

Vision-language models such as VisualBERT, ViLBERT, and LXMERT \cite{li2019visualbert, lu2019vilbert, tan2019lxmert} take object proposals from different vision models as input tokens. This is similar to sViT, where we use a segmentation model for tokenization. The difference lies in the fact that vision-language models input cropped bounding boxes of objects from the image, meaning they do not semantically segment background pixels from objects. Consequently, tokens for these models usually consist of two or more semantic segments as input, increasing the space to explore during training. Furthermore, when using gradient flow to interpret which token is important, it becomes challenging to distinguish whether the object or the background are important. Another significant difference is that these models discard all the pixels that have not been included in the object proposals. However, including these background pixels can be essential for understanding images. For example, background pixels could help the model grasp the context of what is happening with objects. sViT segments objects from the background and does not discard any pixels from an image; instead, it includes those pixels in background tokens.

Data augmentation has garnered significant attention for improving generalization in image models. To increase the diversity of augmented images from a single image, various techniques, such as RandAugment and AutoAugment \cite{cubuk2020randaugment, cubuk2019autoaugment}, were proposed. These methods focus on applying different global augmentation techniques at each training iteration. As mentioned in \cref{sec:aug}, the advantage of our new method of semantic augmentation is that it allows the application of any augmentation technique, and it increases diversity for most techniques compared to when applied to the entire image, so RandAugment and AutoAugment would also benefit from our new method by applying augmentation at the semantic segment level instead of just to the entire image. 

Grad-CAM has found extensive use in interpreting vision models. However, the portion of the images highlighted by Grad-CAM can be hard to interpret due to their lack of localization and semantic meaning. To address some of these issues, ProtoPNet was introduced \cite{chen2019looks}. This model learns patches representing prototypical parts of classes and employs these learned prototypical parts for prediction and interpretation. Yet, as noted by Hoffmann et al. \cite{hoffmann2021looks}, the fact that two patches appear similar to the ProtoPNet model does not necessarily mean they will appear similar to human eyes. ProtoPNet then shares a similar limitation with Grad-CAM. 

\section{Limitation}

Our method has computational limitations compared to ViT. Compared to ViT, sViT requires additional computation for a one-time preprocessing segmentation of the dataset, equivalent to an extra training epoch if segmentation model and training model are similarly sized. However, during inference, sViT doubles the computational cost compared to ViT because it performs segmentation in addition to inference. This highlights a clear trade-off: while ViT is more efficient, sViT boasts greater robustness to out-of-distribution scenarios and distribution shifts, as well as enhanced interpretability. Advancing efficiency in segmentation models could potentially bridge this gap, which is a promising direction for future research.


\section{Conclusion}
\label{sec:conclusion}
We introduce the segmented Vision Transformer (sViT), a vision transformer that employs a semantic segmentation model as a tokenizer. Unlike the original tokenizer, which divides an image into patches without considering semantic information, our model utilizes a segmentation model to separate images into distinct semantic segments or objects, thereby offering higher semantic content. Additionally, we have defined positional and scale embeddings to enhance the model's comprehension of the relative positions of relevant segments and to facilitate rich and efficient data augmentation. Our study demonstrates that the proposed sViT outperforms the original models on non-object-centric datasets, particularly for small datasets. Additionally, it shows a significant accuracy improvement on object-centric datasets, indicating strong generalization in out-of-distribution datasets. We also introduce a new data augmentation paradigm that applies augmentations at the semantic segment level rather than to the entire image. Furthermore, we propose a novel augmentation method to help the model generalize to changes in the relative positions and sizes of the semantic segments. We have shown that this augmentation further improves accuracy across all datasets. Finally, we address the limitations of Grad-CAM and its variants, noting that some highlighted parts are not readily explainable, which impacts interpretability. We discuss how our method overcomes these limitations and provides interpretability in a natural fashion. We believe this work is the first one to explore the use of a semantic segmentation model as a tokenizer, and we hope it will pave the way for further research.

\section*{Acknowledgments}
Work partially supported by ONR, NGA, NSF, Simons Foundation, and Apple.

%
%
\bibliographystyle{splncs04}
\bibliography{main}
\end{document}


\title{Vision Transformers with Natural Language Semantics} 

\titlerunning{Vision Transformers with Natural Language Semantics}
\author{
Young Kyung Kim$^{1}$,
J. Mat\'ias Di Martino$^{1}$,
Guillermo Sapiro$^{1,2}$\\
[2mm]
$^1$~Duke University  \quad $^2$~Apple 
}
\author{Young Kyung Kim\inst{1}\orcidlink{0000-0002-8769-6092} \and
J. Mat\'ias Di Martino\inst{1}\orcidlink{0000-0002-4679-8666} \and
Guillermo Sapiro\inst{1,2}\orcidlink{0000-0001-9190-6964}}

\authorrunning{Y.~Kim et al.}

\institute{Duke University, Durham, NC, 27708, USA
\email{\{yk206,matias.di.martino,guillermo.sapiro\}@duke.edu}\and
Apple, Cupertino, CA, 95014, USA}

\maketitle

\section{Datasets Descriptions}
\label{sec:data}
Places365 (Places) contains $365$ classes and $1.8$ million images; MIT Indoor Scenes (MIT 67) contains $67$ classes and $6700$ images; SUN397 (SUN) consists of ten training and testing partitions, each partition having $397$ classes and $39700$ images; FGCV Aircraft (Aircraft) contains $102$ classes and $10000$ images; Stanford Cars (Cars) contains $196$ classes and $16186$ images; Caltech 101 (Caltech) contains $101$ classes and $9145$ images; ImageNet (INet) contains $1000$ classes and $1.2$ million images; Describable Textures Dataset (DTD) contains $67$ classes $5640$ images; Flower Dataset (Flower) contains $102$ classes and $7169$ images; Food 101 (Food) contains $101$ classes and $101000$ images; and Pet Dataset (Pet) contains $37$ classes and $7349$ images.

ImageNet V2 (INet-V2) contains $1000$ classes and $10000$ images; ImageNet-A (INet-A) contains $200$ classes and $7500$ images; ImageNet-Renditions (INet-R) contains $200$ classes and $7500$ images; and ImageNet-Sketch (INet-S) contains $1000$ classes and $50889$ images. INet-V2 is a collection of images that were gathered following similar protocols as the original ImageNet but during a different time period. INet-A consists of images specifically selected for their adversarial qualities against a vision model trained on the original INet. INet-R includes artistic renditions of various classes from ImageNet. INet-S is a compilation of sketches representing classes from ImageNet.

\section{Boxplot on Number of Tokens for Each Datasets}
\label{sec:boc}

\begin{figure}[t]
  \centering
  \includegraphics[width=\linewidth]{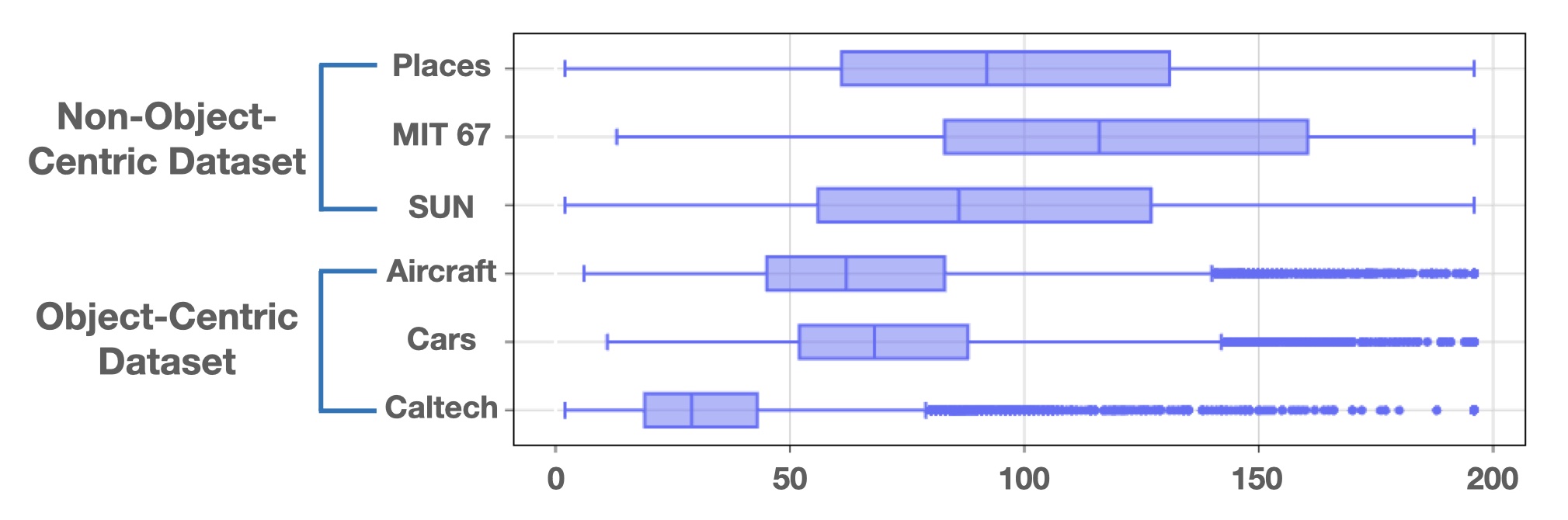}
   \caption{Boxplot of the number of tokens for each dataset. The first three rows represent non-object-centric datasets, while the last three rows are from object-centric datasets.}
   \label{fig:distribution}
\end{figure}

\cref{fig:distribution} in this supplement illustrates the distribution of the number of tokens for each dataset, using SAM as the tokenizer. As we can observe, non-object-centric datasets have a greater number of tokens compared to object-centric datasets. With a higher token count, non-object-centric datasets may provide more signals for learning the contextual information among the semantic segments of images.

\section{Experiment Details}
\label{sec:exp}
For the ViT model, we applied augmentation using the cropping and resizing technique with scale parameters ranging from $0.08$ to $1$ and ratio parameters ranging from $0.75$ to $1.33$. Additionally, a horizontal flip with a probability of $0.5$ was used. Regarding augmentation for sViT, we employed a maximum percentage of $25\%$ for sampling to augment, and cropping and resizing techniques with scale parameters ranging from $0.9$ to $1$ and ratio parameters ranging from $0.75$ to $1.33$. A horizontal flip with a probability of $0.5$ was applied, along with the addition of Gaussian noise to the position and size input, using a parameter with zero mean and a variance of $0.001$. Across all models, we used the Adam optimizer \cite{kingma2014adam}, with $\beta_{1} = 0.90$ and $\beta_{2} = 0.99$, a weight decay of $0.0001$, a batch size of $2048$, and training over 20 epochs on the Places dataset and its subset. For fine-tuning, we employed a linear evaluation protocol, in which only the linear classifier is updated while all other weights in the transformer are frozen. During fine-tuning on the MIT67 and SUN datasets, we preserved the same parameters as in pretraining, trained for an additional 20 epochs, and adjusted the batch size to 512. During fine-tuning on object-centric datasets, except for INet, we trained the sViT for an additional 40 epochs, and the ViT for 60 epochs. For fine-tuning INet, we trained the sViT and ViT for 10 epochs. The experiments were conducted on four A100 GPUs for pretraining, and four A5000 GPUs for fine-tuning. Pretraining on the full Places dataset was completed in approximately 8 hours.

\section{Segmentation and Tokenization}
\label{sec:seg}

\cref{alg:seg_token} summarizes the segmentation and tokenization processes for sViT. 

\begin{algorithm}[h]
   \caption{Segmentation and tokenization}
   \label{alg:seg_token}
\begin{algorithmic}
    \STATE {{\bfseries Output:} List of token embeddings: $list\_emb$}
   \STATE {{\bfseries Input:} Image: $X$, Convolutional layer: $conv$, Neural network: $nn$, Resize function: $resize$, Segmentation model: $seg$}
   \\\hrulefill
   \STATE {$list\_emb =  \{\}$}
   \STATE {$list\_segments, list\_boundbox, list\_size = seg(X)$}
   \STATE {$n\_seg = len(list\_segments)$}
   \FOR{$i=1$ {\bfseries to} $n\_seg$}
   \STATE {$resized\_seg = resize(list\_segments[i])$}
   \STATE {$seg\_emb = conv(resized\_seg)$}
   \STATE {$pos\_emb = nn(list\_boundbox[i],list\_size[i])$}
   \STATE {$emb = seg\_emb + pos\_emb$}
   \STATE {$list\_emb.append(emb)$}
   \ENDFOR
   \STATE {Return $list\_emb$}
\end{algorithmic}
\end{algorithm}

\section{Additional Example for Interpretable results}
\label{sec:int_appendix}

\begin{figure*}[t]
  \centering
  \includegraphics[width=0.89\linewidth]{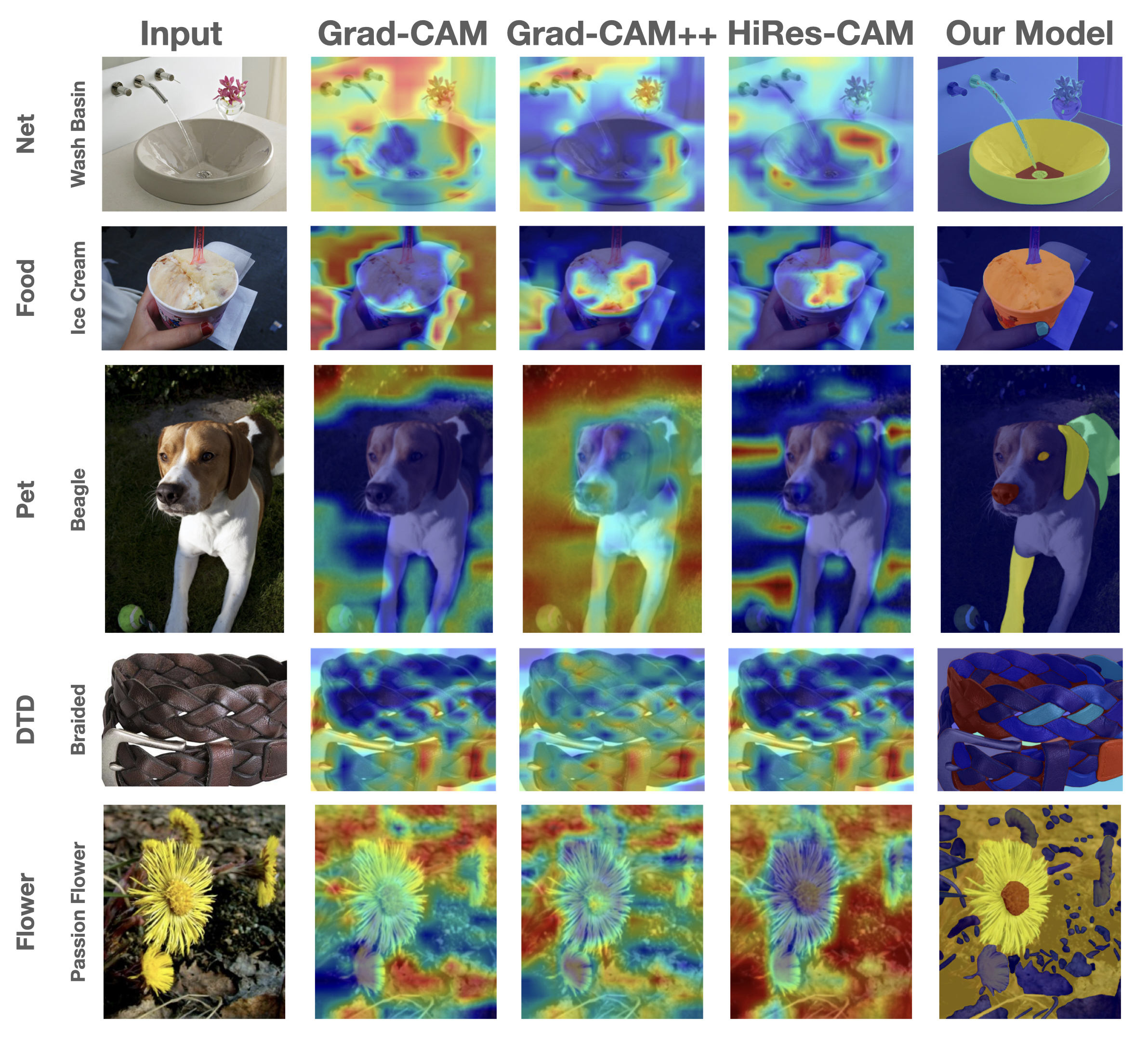}
   \caption{The first column shows the images used to evaluate the interpretability of ViT and sViT. Columns two to four represent the interpretable outcomes from applying Grad-CAM, Grad-CAM++, and HiRes-CAM on ViT, respectively. The fifth column shows the interpretable outcome for our method. The color coding indicates the level of importance. The importance escalates progressively through blue, green, yellow, and orange before reaching the peak at red. Only our proposed method consistently provides humanly interpretable and semantic results.}
   \label{fig:interpret_appendix}
\end{figure*}

\cref{fig:interpret_appendix} compares the interpretability of Grad-CAM, Grad-CAM++, HiRes-CAM, and our new methods across the remaining datasets.

\section{Ablations studies on segment-level augmentation}
\label{sec:seg}

In our ablation studies, we experimented with applying various segment-level augmentations to assess their impact on performance. These augmentations included horizontal flip (flip), crop and resize (crop), and the addition of Gaussian noise into the positional embedding (pos). We also explored combinations of these augmentations to examine their cumulative effects on performance, as presented in \cref{tab:abl}.

\begin{table}
\caption{Evaluation of different segment-level augmentation on sViT$16$, and sViT$32$ models. }
  \centering
  \begin{tabular}{lccccccr}
    \toprule
    Method & Aircraft          & Cars &    Caltech & MIT67  & SUN \\
    \midrule
    sViT$16$ & 76.99 &	72.9 &	76.03 &	68.95& 50.77  \\
    sViT$16$ flip & 78.4 &	73.12 &	77.13 &	69.62 &	50.88  \\
    sViT$16$ crop & 77.41 &	72.51 &	77.47 &	69.32 &	50.65  \\
    sViT$16$ pos & 74.47 &	72.13 &	76.54 &	70.07 &	50.97  \\
    sViT$16$ flip+crop & 77.71 &	72.76 &	77.69 &	69.25 &	50.68  \\
    sViT$16$ combined & 77.01 &	72.46 &	76.36 &	70.75 &	51.03  \\
    \midrule
    sViT$32$ & 79.97 &	71.85 &	76.85 &	70.59 &	51.06  \\
    sViT$32$ flip & 80.59 &	73.92 &	76.87 &	70.45 &	51.47  \\
    sViT$32$ crop & 78.7 &	72.62 &	76.22 &	70.67 &	51.38  \\
    sViT$32$ pos & 78.64 &	72.78 &	75.23 &	71.34 &	51.05  \\
    sViT$32$ flip+crop & 78.67 &	73.59 &	76.84 &	71.37 &	51.26  \\
    sViT$32$ combined & 79.23 &	73.6 & 	77.07 &	72.01 &	51.53  \\
    \bottomrule
  \end{tabular}
  \label{tab:abl}
\end{table}

Interestingly, adding Gaussian noise to the positional embedding did not enhance performance in most object-centric and fine-grained datasets. However, we observed a noticeable improvement in scenery datasets. This disparity suggests that the relative positions of segments are particularly crucial in scenery images, where the spatial arrangement plays a pivotal role in defining the scene's context. Conversely, for object-centric and fine-grained datasets, where the focus is more on individual elements rather than their spatial relationships, this type of augmentation appears less effective.

\section{Interpretable `mistakes'}
\label{sec:int_mis}
\begin{figure}[t]
  \centering
   \includegraphics[width=0.94\linewidth]{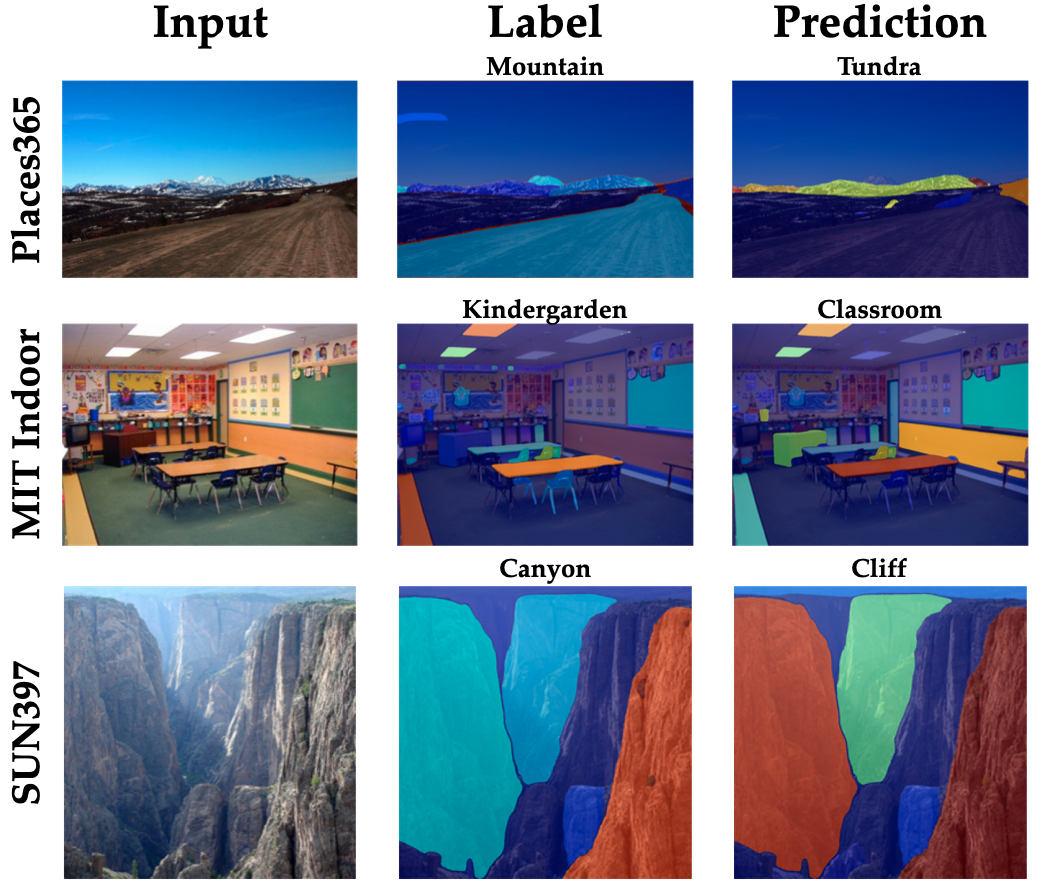}
   \caption{sViT provides interpretable `mistakes' The first column shows the input image for sViT. Column two represents the interpretable outcome from the actual label. Column three represents the interpretable outcome from the model's prediction. The color coding indicates the level of importance. The importance escalates progressively through blue, green, yellow, and orange before reaching the peak at red. sViT clearly shows the reasonable mistake of the system, which upon examination of the model's interpretability, are not necessarily mistakes.}
   \label{fig:interpret_mis}
\end{figure}

The effectiveness of our method is further illustrated in \cref{fig:interpret_mis} of this supplement, particularly in our ability to comprehend why the model sometimes generates inaccurate predictions. Even when the sViT model makes errors in its predictions, our approach allows us to discern the reasons behind such mistakes. For instance, consider the scenario where sViT predicts `tundra' for an input image labeled as `mountain.' Upon observing the highlighted region, it becomes apparent why the model made this prediction, as it emphasizes a snowy mountain, a characteristic often found in tundra landscapes. Similar insights are evident in other examples (\cref{fig:interpret_mis}). In specific instances, the model inaccurately predicts a `kindergarten' image as a `classroom' or a `canyon' image as a `cliff,' which, upon examination of the highlighted semantic segments, appears reasonable since such objects are commonly found in classroom or cliff images. 
\bibliographystyle{splncs04}
\bibliography{main}